\documentclass[review]{elsarticle}
\usepackage{hyperref}
\usepackage{amssymb}
\usepackage{csquotes}
\usepackage{booktabs}
\usepackage{dirtytalk}
\usepackage{soul}
\usepackage{graphicx}
\usepackage{amssymb}
\usepackage{algorithm}
\usepackage{algorithmic}
\usepackage{csquotes}
\usepackage{tikz}
\usepackage{amsmath}
\usepackage{listings}
\usepackage{graphicx}
\usepackage{mathtools}
\mathtoolsset{centercolon}
\newtheorem{theorem}{Definition}
\journal{Journal}
\begin{document}
\begin{frontmatter}
\title{Declarative Memory-based Structure for the Representation of Text Data}
\author[cuj,nit]{Sumant Pushp\corref{cor1}}
\ead{sumantpushp@gmail.com}
\author[nit]{Pragya Kashmira}
\author[iitg]{Shyamanta M Hazarika}
\cortext[cor1]{Corresponding author}
\address[cuj]{Department of Computer Science and Technology, Central University of Jharkhand, Ranchi, India}
\address[nit]{Department of Computer Science and Engineering, National Institute of Technology, New Delhi, India}
\address[iitg]{Department of Mechanical Engineering, Indian Institute of Technology Guwahati, Assam, India}
\begin{abstract}
In the era of intelligent computing, computational progress in text processing is an essential consideration. Many systems have been developed to process text over different languages. Though, there is considerable development, they still lack in understanding of the text, i.e., instead of keeping text as knowledge, many treat text as a data. In this work we introduce a text representation scheme which is influenced by human memory infrastructure. Since texts are declarative in nature, a structural organization would foster efficient computation over text. We exploit long term episodic memory to keep text information observed over time. This not only keep fragments of text in an organized fashion but also reduces redundancy and stores the temporal relation among them. Wordnet has been used to imitate semantic memory, which works at word level to facilitate the understanding about individual words within text. Experimental results of various operation performed over episodic memory and growth of knowledge infrastructure over time is reported.
\end{abstract}
\begin{keyword}
Cognitive Psychology \sep Knowledge Representation \sep Episodic Memory \sep Text Processing
\MSC[2010] 00-01\sep  99-00
\end{keyword}
\end{frontmatter}

\section{Introduction}
Knowledge could be identified as a ingredient for the computational process of thinking, irrespective of whether  the computation is rational or irrational. A rational computation yield an intelligent behaviour by making best use of knowledge whereas an unintelligent behaviour could be well understood as; %

\begin{displayquote}
\textit{When we say that someone has behaved unintelligently, like when someone has used a lit match to see if there is any gas in a car's gas tank, what we usually mean is not that there is something that person did not know, but rather that person has failed to use what he or she did know.}
\end{displayquote}
\rightline{{\textit{--- Brachman and Levesque}\cite{brachman1992knowledge}}}
\par
Vast knowledge is available to us since the advent of life. But, all this knowledge is mere useless until it can be utilized so as to be able to determine its implication and draw relevant inference from it\cite{mccarthy2008well}. Evidences\cite{hodges1999and,pushp2018cognitively} suggest that in order to do so, human brain structure the information. A multi store model of human memory system\cite{atkinson1968human} was proposed, way back in 1968. The model claims a memory system specifies the underlying infrastructure for rational behaviour for animals. Human intelligence immensely rely on such infrastructure. Among all such type of memory structure Long Term Memory(LTM) is of special interest as it can store information for unlimited period of time. Further Declarative and non declarative are two logical separation of LTM,  where first can store knowledge that someone can tell others and later contains knowledge that someone can show by doing\cite{anderson2004integrated,saikia2016cbdi}. Two different type of declarative memory - semantic memory and episodic memory keeps general world knowledge and experience gained over time respectively. Briefly, memory includes those aspects of human which they gain by their own observation and experiences which changes over time to revise beliefs about an object, event, or action of the world. 
\par
Research in the field of knowledge representation has been pursued since the concept of modern computer. There  has been much development in the area over the last few decades\cite{davis2015commonsense}. In the context of intelligent processing, most of the existing knowledge representation techniques provide strong evidence towards processing of a particular language for a particular problem, but most of the such system lack the significant objective, of \textit{"generalization"}.
\par
Text processing usually involves performing various computations on the unstructured text. A more efficient way to process text could be obtained by first transforming the text into a structured format and then fed for processing. There is growing interest for knowledge representation technique for natural language\cite{davis2015commonsense} with systems that are able to store information in a structured format so that we could say it \textit{knowledge}. 
\par The remainder of this paper is organized as follows: section~\ref{backandmo} provides the background review and motivation towards the representation followed by desidarata in  Section~\ref{desidarata}. Section~\ref{formalstruct} contains the formal structure to illustrate the fundamental intuition of representation followed by experiment and evaluation in section~\ref{evaluation}. Finally section~\ref{conclusion} briefly concludes the paper.
\section{Background and Motivation}
\label{backandmo}
Human memory has long been studied and attempts have been made to map the  same,  but it is a very complex system whose absolute structure hasn't yet been obtained. Recent research \cite{radvansky2015human} has shown that human memory is not located in one particular place in the brain, but is instead a distributed structure in which different parts of the brain act in co-ordination with one another. Thus, an actual representation might be a large complex network, in which the nodes symbolize the various elements that join at edges to form a memory. 
A memory system specifies the underlying infrastructure for rational behaviour for animals. Human intelligence immensely rely on such infrastructure. Briefly, memory includes those aspects of human knowledge which they gain by their own observation and experiences which changes over time to revamp beliefs about objects, events or actions of the world. 
\par
Psychological studies fundamentally classify human memory system in two classes based on their temporal existence 1. Short Term Memory(STM) and 2. Long Term Memory(LTM). Short Term Memory holds current observation and survive for very short periods of times, which are assumed to be in the order of seconds\cite{laird1}. Where as Long Term Memory can detain acquired knowledge for unlimited duration\cite{anderson}. LTM is further divided into two subcategory\cite{atkinson1968human}; Declarative and Non-declarative storage, where first can store knowledge that someone can tell others and later contains knowledge that someone can show by doing\cite{baddeley1997human}, which is of our special interest. Evidences suggest that declarative memory contains two different types of declarative knowledge, which are separately stored in different storage infrastructure known as semantic memory and episodic memory\cite{anderson}. 
\par
\textit{Semantic memory} refers to the facts, information and features about objects which are internally used by brain in order to determine what the object is. Semantic Memory is used in the field of AI in order to determine the meaning of a word or fact such that the computer system is able to understand it and perform computations on it. \textit{Episodic memory} represents the chronological record of a persons experience, where some specific events are only stored for long interval of time\cite{bliss1993synaptic}. Episodic Memory can be explicitly accessed and an episode can be reconstructed using it. 
When the personal context is shared from the Episodic Memory, it becomes a part of the Semantic Memory of the person. This generally happens when there are some facts or information in the episodic memory which is repeatedly learnt over time \cite{mckoon1986critical,greenberg2010interdependence}. 
\par
Beyond representation, it also important to define how operations are performed on knowledge. While retrieving instances of knowledge it is important to determine which information is relevant to the task at hand. Insignificant knowledge needs to be removed so as to avoid accumulating non-essential information. The essence is reinforce important parts to make sure that the knowledge is not forgotten. Five different operations  that a knowledge representation technique should be able to perform in order to enable the aforesaid are listed below\cite{nuxoll2007enhancing}: 
\begin{enumerate}
    \item \textit{Encoding} deals with determining how a new data will be transformed following the rules of the KR such that the original structure of the KR is maintained. Also, it has to be taken care of when encoding would be initiated. 
    \item \textit{Storage}  handles the internal storage  structure for the KR. Further, it needs to ascertain how the dynamics of the storage will change when any other operation is performed, so as to reflect the operation but at the same time retain the structural rules.
    \item \textit{Retrieval} manages when and how retrieval will be triggered. Most importantly, the process involved in the retrieval should be defined in detail, which involves initiation condition, selection methodology, and similarity determination.
    \item \textit{Forgetting} tackle the removal of insignificant knowledge so as to avoid accumulating non-essential information. 
    \item \textit{Consolidation} take care of reinforcing important information to make sure that knowledge is not forgotten while it is in use. Further, care must be taken such that knowledge which is used frequently becomes a permanent memory after reaching a frequency threshold. 
\end{enumerate}
The essential requirement for any system to exhibit human-like intelligence is to be able to draw conclusions from the knowledge the system already possesses. This requires the system to be able to represent relationships between various beliefs, includes not only inferring any new rules encountered but also to be able to update the existing ones. In order to establish an intelligent conversation, the system must be capable of determining the feasible choices by associating it with the existing patterns and then on the basis of the feasibility of various choices, be able to choose a alternative. 
\section{Desiderata}
\label{desidarata}
The human-machine interaction system can work towards achieving the goal only when it supports some varied capabilities that are required for a system to achieve human level intelligence \cite{pat,pushp2017cognitive}. Desiderata for completeness and evaluation of a cognitive agent for human machine communication, whether directly or through embedded process are:
\par
Adaption could be bolstered if one can properly categorize knowledge and can recognize and extract rational knowledge. Therefore a proper knowledge structure along with satisfactory retrieval mechanism is essential.
\par
The essential requirement for any system to exhibit human-like intelligence is to be able to draw conclusions from the knowledge the system already possesses. This requires the system to be able to represent relationships between various beliefs, which includes not only inferring any new rules encountered but also to be able to update the existing rules.
\par
The system must possess the ability to encode and store the result of previous experiences and to be able to retrieve them later and the inferences drawn by them at the previous stage. Additionally, the rules must be generalized in memory to be able to learn by applying them to similar problems or other tasks in the same domain. 
\par
Aforementioned desiderata express the abstract notion of underlying requirement. Throughout the work we have emphasize on developing the knowledge infrastructure which is fulfilling part of requirement discussed above. The proposed mechanism is psychologically plausible and never denies the possibility of other better methodology. The design perspective would go around the human centered functionality.
\section{Episodic Memory and Knowledge Representation}
\label{formalstruct}
Knowledge representation fundamentally is a function which takes input from one domain and returns an output belonging to another one. In other words, it is a mapping of entity between two domain. There could be enormous variety of structural method through which one could perform such a mapping. Due to human centric bias, this section will supply the formal description of episodic memory based knowledge representation structure for text data. 
\par 
Episodic memory in general is a network of experience\cite{anderson} gained by individual. Which is certainly different for different individual. Since experiences are gained over time, therefore it requires a temporal order of maintenance. Multiple experiences gained over time could have variety of interconnected contexts, that is again a crucial challenge for modeling.

\begin{theorem}
Episodic memory is a 5 tuple consists of chronological sequence of episodes and temporal relationship between them.
\end{theorem}
\begin{align*}
    E_{memory}
    \vcentcolon =
    (
    ep_{id},
    ep_{timestamp},
    ep_{next},
    U_{tility},
    L_{ink}
    )
\end{align*} 
where, $ep_{id}$ refers to the unique identification associated with individual episode,
$ep_{timestamp}$ records the temporal parameter for each instance of individual episodes, which further could be used to establish the temporal occurrence of individual episodes. $ep_{next}$ indicate the references to nextepisode; whereas significance of a particular instance of episode could be recorded in $U_{tility}$ tuple.
$L_{ink}$ establishes the similarity connection between instances of episode which itself composed of two tuple $< N_{id}$,$l_{weight} >$. Pair of parameter defines the strength of similarity between episodes. It comes into picture when a particular episode is linked with number of other episodes due to context similarity.
\begin{figure*}[h!]
\centering
\includegraphics[scale=0.20]{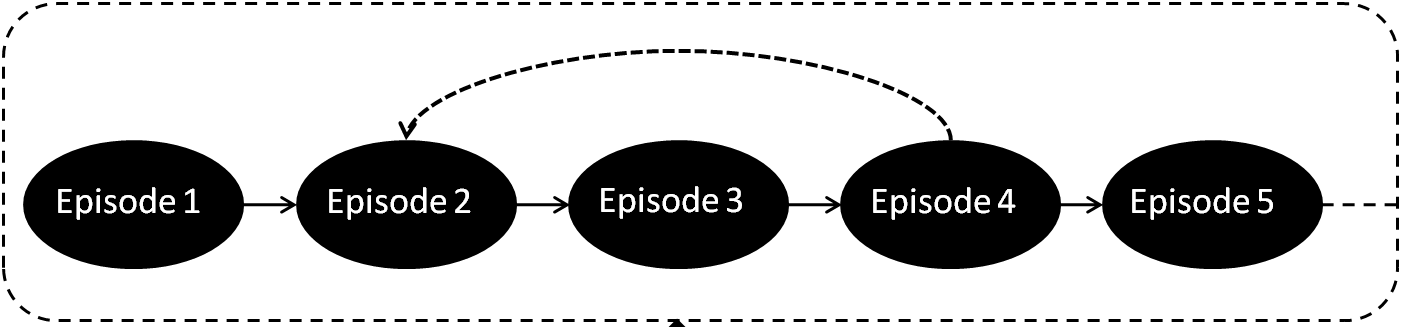}
\caption{Sequence of Episodes in Episodic Memory}
\label{fig:episode_sequence}
\end{figure*}
\par The knowledge representation generates a graph where there are various episodes linked in chronological order, as shown in Figure ~\ref{fig:episode_sequence}. In the memory it can serve to find the last gained knowledge about a particular context as well as to answer a query whose context could be inferred from previous experience.
\par Episodic memory is composed of instances of episodes linked together depending on the strength of similarity and temporal occurrence. Among such instances, a few could be more significant then other in a particular context. 
\begin{theorem}
Each episode in turn consists of instances which have smaller data chunks stored in the form of nodes. Instances are connected to previous instances on the basis of similarity. An episode from node s to g is a sequence of nodes ($n_0$, $n_1$, $n_2$,...,$n_k$) such that s= $n_0$ , g= $n_k$. Individual episode could be formally expressed as a 5-tuple entity;
\end{theorem}
\begin{align*}
    E_{pisode}
    \vcentcolon =
    (
    ep_{instanceID},
    N_{odes},
    T_{imestamp},
    N_{ext},
    U_{tility}
    )
\end{align*}
where, $ep_{instanceID}$ is again unique identification for individual nodes within an instance of an episode. 
$N_{odes}$ keep the sequence information about the nodes i.e. the order of occurrence of nodes within an instance of episode. $T_{imestamp}$ indicate the temporal parameter about individual nodes within each instance of episode.  $N_{ext}$ contains the reference to next node within an episode. $U_{tility}$ is again there to state the significance or importance about the context of particular node. This would be initially fixed constant value for any newborn episode and can change according to uses or priority of the episode.
\par Individual episodes are further composed of series of nodes; which are organized as a collection of sub nodes expanded in three logical layers; primary, secondary and ternary. 
\begin{theorem}
A node is an elementary data unit of episodic memory. It consists of vital decision making information. A node is a 5 tuple structure.
\begin{align*}
    N_{ode}  
    \vcentcolon =
    (
    N_{ID},
    D_{ata},
    T_{ag},
    N_{episodeID},
    T_{ype}
    )
\end{align*}
\end{theorem}
\par Where, $N_{ID}$ indicates the unique identification of a node. Depending on type of node (indicated by $T_{ype}$ ), a node contains relevant information; $D_{ata}$ refers to the collection of keywork associated with a node in a context, $N_{episodeID}$ records the episode identification of within which current node belongs.
\par
As an elementary data unit and in line with the complexity of text representation, the three types of nodes are a logical separation of text is based on the building block of a particular language, which turns out to be a language dependent functionality.
\par
Primary node indicates the subject of the sentence. It acts as an anchor to the underlying context. Such that it could be used as a reference point whenever that is called upon in future. The primary node has been chosen from the tags such that it contains the information regarding the main subject of the instance. Table~\ref{primary} contains the list of tags associated with primary node. 
\begin{table}[htb!]
\centering
\resizebox{0.4\textwidth}{!}
{
\begin{tabular}{c c} 
 \hline
 Tag & Meaning \\ 
 \hline
NN & Noun, singular or mass\\
NNS & Noun, plural\\
NNP & Proper noun, singular\\
NNPS & Proper noun, plural\\
\hline
\end{tabular}
}
\caption{Tags for Primary Node}
\label{primary}
\end{table}
\par 
Secondary node works as the sub nodes of primary and keeps the information about the subject of the primary node. This helps in determining the attributes of the subject. Table~\ref{secondary} comprises the possible tags to identify a secondary node.
\begin{table}[htb!]
\centering
\resizebox{0.6\textwidth}{!}
{
\begin{tabular}{c c} 
 \hline
Tag & Meaning\\
\hline
VB & Verb, base form\\
VBD & Verb, past tense\\
VBG & Verb, gerund or present participle\\
VBN & Verb, past participle\\
VBP & Verb, non-3rd person singular present\\
VBZ & Verb, 3rd person singular present\\
JJ & Adjective\\
MD & Modal\\
\hline
\end{tabular}
}
\caption{Tags for Secondary Node}
\label{secondary}
\end{table}
\par 
Further, tertiary node indicate the adverbs related to the property being referred to in the secondary node. Table~\ref{tertiary} shows two observed tag for tertiary node.
\begin{table}[htb!]
\centering
\resizebox{0.5\textwidth}{!}
{
\begin{tabular}{ c c c } 
 \hline
Tag & Meaning & Secondary\\
\hline
RB & Adverb & JJ\\
RBS & Adverb, Superlative & JJ\\
\hline
\end{tabular}
}
\caption{Tags for Tertiary Node}
\label{tertiary}
\end{table}

\subsection{Operations}
Beyond primitive structure, we present how variety of operations would operate over the dynamic structure to deal with the transformation of informal input to a computable knowledge structure. While retrieving instances of knowledge it is also important to determine which information is relevant to the task at hand. Also, it is important that the insignificant knowledge be removed so as to avoid accumulating non-essential information. At the same time, it is also of essence to reinforce important parts to make sure that the knowledge is not forgotten. Various operations \cite{nuxoll2007enhancing} that a knowledge representation technique should be able to perform in order to enable the aforesaid are: 
\subsubsection{Encoding} Encoding deals with determining how a new data will be transformed following the rules of the Knowledge Representation such that the original structure of the \textit{Knowledge} is maintained. 
\par
Whenever a new action sequence is observed, it is required to examine whether the new action sequence would be considered as a new episode or an instance of the ongoing episode. We assert in general that if elapsed time between timestamp of current instance and timestamp value of last observed instance is greater then the time elapsed between the last observed instance and first instance of that episode then new action sequence would be considered as start of new episode. 
\begin{theorem}
Episode determination is assimilated as a Boolean valued function. Function $\delta_{e}$ returns 1 if new episode is required to be started, similarly returning 0 indicate that current instance continues with the current episode. We would consider $\varphi_{f}$, $\varphi_{l}$ and $\varphi_{c}$ as different timestamp $\varphi_{f}$ and $\varphi_{l}$ indicates start timestamp and last timestamp of episode in which last instance belongs. Further $\varphi_{c}$ returns the timestamp of current instance for which we would like to determine the episode.
\[\delta_{e} =
  \begin{cases}
    1                           & if~(\varphi_{c}-\varphi_{l}) \geq ~\tau *(\varphi_{l}-\varphi_{f}) \\
    0                           & Otherwise\\
  \end{cases}
\]
Where $\tau$ is the time-stamp constant which is a tuning factor that may be set according to the application requirement i.e. when an application demands new time-stamp be created within a small time-interval, smaller value of $\tau$ would be preferable.
\end{theorem}
\par
Nodes are an elementary unit of proposed episodic memory which is formed based on the classification of input text based on a tag it acquire. Tag features of the input text are used to organize the node as a primary, secondary and ternary node.

\subsubsection{Storage} Storage handles the internal storage  structure for the KR. The storage handles the changes when any other operation is performed, so as to reflect the operation but at the same time retain the structural rules.
\par
The structure used to store an episode will influence the efficiency of addition or modification of knowledge as well as relevant retrieval. As it has been experimentally established that graphs would best complement to assimilate a non linear structure like an episodic memory\cite{nuxoll2007enhancing}. Therefore, to take care of such structure it maintains an interaction graph. 
\begin{theorem}
The Interaction graph is a directed graph denoted as $G_I$, is of two tuple structure of ($V_E$,$L_E$).  Where $V_E$ is a finite, nonempty set of episodes and $L_E$ is a set of links between pair of episodes. 
\end{theorem}
Conventionally, the set E and L represents the vertices and edges of a graph respectively.
\par
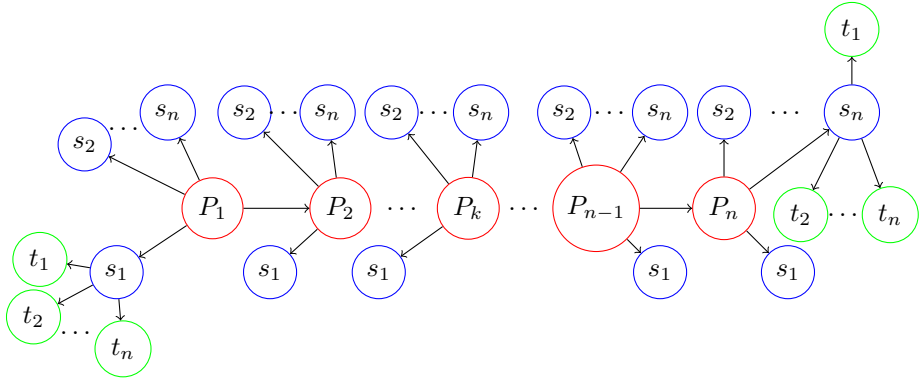
\begin{figure*}[htb!]
    \centering
\begin{tikzpicture}[scale=0.85]
    \node[shape=circle,draw=red] (p1) at (0,0) {$P_1$};
    \node[shape=circle,draw=red] (p2) at (2,0) {$P_2$};
    \node[shape=circle,draw=red] (p3) at (4,0) {$P_k$};
    \node[shape=circle,draw=red] (pn-1) at (6,0) {$P_{n-1}$};
    \node[shape=circle,draw=red] (pn) at (8,0) {$P_{n}$};
    
    \path [->](p1) edge node {} (p2);
    \path (p2) to node {\dots} (p3);
    \path [->](pn-1) edge node {} (pn);
    \path (p3) to node {\dots} (pn-1);
    
    \node[shape=circle,draw=blue] (p1s1) at (-1.5,-1) {$s_1$};
    \node[shape=circle,draw=blue] (p1s2) at (-2,1.0) {$s_2$};
    \node[shape=circle,draw=blue] (p1sn) at (-0.7,1.5) {$s_n$};
    \path [->](p1) edge node {} (p1s1);
    \path [->](p1) edge node {} (p1s2);
    \path [->](p1) edge node {} (p1sn);
    \path (p1s2) to node {\dots} (p1sn);
    
    \node[shape=circle,draw=blue] (p2s1) at (0.9,-1) {$s_1$};
    \node[shape=circle,draw=blue] (p2s2) at (0.5,1.5) {$s_2$};
    \node[shape=circle,draw=blue] (p2sn) at (1.8,1.5) {$s_n$};
    \path [->](p2) edge node {} (p2s1);
    \path [->](p2) edge node {} (p2s2);
    \path [->](p2) edge node {} (p2sn);
    \path (p2s2) to node {\dots} (p2sn);
    
    \node[shape=circle,draw=blue] (p3s1) at (2.6,-1) {$s_1$};
    \node[shape=circle,draw=blue] (p3s2) at (2.8,1.5) {$s_2$};
    \node[shape=circle,draw=blue] (p3sn) at (4.2,1.5) {$s_n$};
    \path [->](p3) edge node {} (p3s1);
    \path [->](p3) edge node {} (p3s2);
    \path [->](p3) edge node {} (p3sn);
    \path (p3s2) to node {\dots} (p3sn);
    
    \node[shape=circle,draw=blue] (pn-1s1) at (7,-1) {$s_1$};
    \node[shape=circle,draw=blue] (pn-1s2) at (5.5,1.5) {$s_2$};
    \node[shape=circle,draw=blue] (pn-1sn) at (7,1.5) {$s_n$};
    \path [->](pn-1) edge node {} (pn-1s1);
    \path [->](pn-1) edge node {} (pn-1s2);
    \path [->](pn-1) edge node {} (pn-1sn);
    \path (pn-1s2) to node {\dots} (pn-1sn);
    
    \node[shape=circle,draw=blue] (pns1) at (9,-1) {$s_1$};
    \node[shape=circle,draw=blue] (pns2) at (8,1.5) {$s_2$};
    \node[shape=circle,draw=blue] (pnsn) at (10,1.5) {$s_n$};
    \path [->](pn) edge node {} (pns1);
    \path [->](pn) edge node {} (pns2);
    \path [->](pn) edge node {} (pnsn);
    \path (pns2) to node {\dots} (pnsn);

   \node[shape=circle,draw=green] (s1t1) at (-2.7 ,-0.8) {$t_1$};
   \node[shape=circle,draw=green] (s1t2) at (-2.8,-1.7) {$t_2$};
   \node[shape=circle,draw=green] (s1tn) at (-1.4,-2.2) {$t_n$};
    \path [->](p1s1) edge node {} (s1t1);
    \path [->](p1s1) edge node {} (s1t2);
    \path [->](p1s1) edge node {} (s1tn);
    \path (s1t2) to node {\dots} (s1tn);

   \node[shape=circle,draw=green] (snt1) at (10 ,2.8) {$t_1$};
   \node[shape=circle,draw=green] (snt2) at (9.2,-0.1) {$t_2$};
   \node[shape=circle,draw=green] (sntn) at (10.6,-0.1) {$t_n$};
    \path [->](pnsn) edge node {} (snt1);
    \path [->](pnsn) edge node {} (snt2);
    \path [->](pnsn) edge node {} (sntn);
    \path (snt2) to node {\dots} (sntn);
\end{tikzpicture}
 \caption{Schematic View of Nodes within an Episode}
 \label{nodes}
\end{figure*}

Encoded tags are used to classify the input text into three node type; primary , secondary and ternary. Storage structure treat primary node of the input text as root for that node instance. Secondary nodes are directly connected to primary node whereas ternary nodes are directly connected to secondary nodes. Therefore a evolutionary model of knowledge structure is stored in form of node for each input instance. A symbolic structure is shown in the figure~\ref{nodes}.
\subsubsection{Retrieval}
Retrieval manages when and how knowledge will be triggered. The retrieval process defines how the past episodes and instances may be retrieved from the storage. It involves initiation condition, selection, and similarity determination
\par
Spontaneous retrieval is initiated when an episode is retrieved in order to link similar instances whereas deliberate retrieval is initiated when the conversation demands an answer which might be present in the memory. The knowledge structure backtracks and finds the first node it encounters that has the same primary node or similar to the current node. Here Wordnet\cite{WinNT} plays significant role to determine the similar knowledge. \begin{theorem}
Whenever it encounters a node which satisfies this condition, it applies the function \textit{S}, where \textit{i} is the node on which function is applied and computes value for both, the current node and the node being compared, \textit{j}, using the equation,
\begin{align*}
    {
    \mathbb{S} = k \cdot \eta_{p} + l \cdot \eta_{s} + m \cdot \eta_{t}
    }
\end{align*}
Where, \textit{k, l, m }are constants such that k + l + m = 1 and \(k > l > m\), in order to ensure that maximum priority is given to the primary node. 
\par
Here, $\eta_{p}$ is the inverse of distance between the primary nodes which is calculated by finding the similarity between them using WordNet, $\eta_{s}$ is the inverse of least distance between the secondary nodes and $\eta_{t}$ is the inverse of least distance between the tertiary nodes.
\end{theorem}
\par
If \(\mathbb{S}>\rho\) (threshold constant) then, the difference between the two nodes is acceptable and they are linked, otherwise previous nodes are searched until similar node is found or start of time is reached.
The weight of any new link will be initialized at the time episode is linked as 1. 
\subsubsection{Forgetting}Forgetting performs the removal of insignificant knowledge to avoid accumulating non-essential information. Therefore the organization of the episodic memory changes over time\cite{nuxoll2010comparing}. Usually forgetting target those episodes which is least used. Therefore it weakens the link and decrease the utility value of that instance over the period of time.
\begin{theorem}
Link weight has to decrease to weaken the significance of an instance. At time $t$, the new weight ${w_t}$ depends on the difference in time elapsed between target instance to the current instance and utility of the target instance. Therefore, ${w_t}$ could be updated as;
\begin{align*}
w_{t} = x * w_{t-1} - y*d_{ci} + z*u_{t}
\end{align*}
Where $d_{ci}$ is the difference between the time stamp of target instance and current one. $u_{t}$ refers to the utility of the particular instance. Further, $x$ is link weight constant, $y$ is forgetting constant and $z$ is utility constant.
\end{theorem}
The constants \textit{x, y, z} are tuning factors whose values may be taken according to the application. Value of \textit{x} can be chosen in the range (0,1) where values close to 1 implies slowest forgetting and closer to 0 implies rapid forgetting. Value of \textit{t} must be chosen such that effect of passage of time may be reflected on the link, where the value must be kept in the range (0,1) with greater value signifying rapid decrease in weight links with time.  The weight of any new link will be initialized at the time episode is linked as 1, while the utility of instance is fixed as 1 at the time of instance creation itself.
\begin{theorem}
The utility value of an instance $u_t$ at time $t$ of an instance will change with respect to the elapsed time measure. Therefore new utility could be computed as,
\begin{align*}
    u_{t} = u_{t-1} - y * d_{ci}
\end{align*}
where, \textit{y} is the forgetting constant which will be consistent with the forgetting value with respect to time of links.
\end{theorem}
\par
If the weight of any link is lower than link threshold, \(\alpha\) or the utility of an instance is lower than utility threshold, \(\beta\) then the link should be severed or the instance be deleted, respectively. The deletion of instances or severing of links must be done while the application is idle.
\subsubsection{Consolidation}Consolidation take care of reinforcing important information to make sure that the knowledge is not forgotten while it is in use. Also, it must be taken care of that some knowledge which is used frequently becomes a permanent memory after reaching a frequency threshold. 
\par
Whenever, difference is acceptable, and the instances are linked, the utility value of both the instances and the weight of the link is increased using the equation,
\begin{align*}
    w_{t} =(1- x) * w_{t-1} - (1-y)*t +(1-z) * u_{t}
\end{align*}
The weight of the link is increased using the same parameters as the link weakened has to be performed during forgetting.
\begin{align*}
    u_{t} = \frac{ 1}{ \rho} *u_{t-1} + y * t
\end{align*}
The utility value at time \textit{t} is increased with reference to its previous value, threshold constant and forgetting constant. Consolidation is performed when the system is idle.
\section{Experiments, Evaluation and Result}
\label{evaluation}
In this section, we present the experiments conducted to observe the working of each operation. At different instance of time, different input paragraphs are introduced to examine the snapshot of episodic memory coupled with various operators. Table~\ref{encode-episode} presents a snapshot of existing episode ahead of introducing input. Creation of episodes, instances of node within individual episode and effectiveness of operators will be observed to examine the working of knowledge structure.Firstly, three different paragraphs are given as input at three separate instance of time and observation will be made thereafter.
\begin{displayquote}
\textit{Input:} \textit{The sun is a huge ball of gases. \\The Sun is mainly made up of hydrogen and helium gas. \\The surface of the Sun is known as the photosphere.}
\end{displayquote}

\begin{table}[h!]
\centering
\resizebox{\textwidth}{!}{%
\begin{tabular}{@{}lllllll@{}}
\hline

\multicolumn{1}{c}{Instance}  
& \multicolumn{2}{c}{Link} 
& \multicolumn{1}{c}{Timestamp}                
& Instance   
& Next 
& Utility 
\\ 
\cline{2-3}
& ID         
& Weight      
&                             
&            
&      
&         \\ \hline
4300673472 & None       & 0           & 2018-12-09 10:09:08. 532000 & 4300679876 & None & 1       \\ \hline
\end{tabular}%
}
\caption{Instance of an Episode before encoding the above said input}
\label{encode-episode}
\end{table}

\par
We would visualize the instances of node correspond to individual sentence in Table~\ref{encode-instance-1}. Individual row in Table~\ref{encode-instance-1} reflects the unique identification mark for individual node, fragments of input belongs to variety of node type, time-stamp and next linked node.  
\begin{table}[h!]
\centering
\resizebox{\textwidth}{!}{%
\begin{tabular}{@{}lllllll@{}}
\hline
ID 
& Primary 
& Secondary
& \multicolumn{2}{c}{Tertiary}
& Timestamp 
& Next 
\\

\cline{4-5}
& 
& 
& 
Prev 
& Tag 
& 
&  
\\
\hline

4300679876 
& ['sun', 'ball', 'gases']  
& ['huge'] 
& [] 
& [] 
& 2018-12-09 10:09:08. 485000 
& 4300679886  
\\
\hline
4300679886 
& ['Sun', 'hydrogen', 'helium', 'gas'] 
& ['made'] 
& [1] 
& ['mainly'] 
& 2018-12-09 10:09:08. 485000  
& 4300679896
\\
\hline
4300679896& ['Sun', 'surface', 'photosphere'] 
& ['known'] 
& [] 
& [] 
& 2018-12-09 10:09:09. 485000 
& None
\\
\hline
\end{tabular}
}
\caption{Instance of Nodes after Encoding}
\label{encode-instance-1}
\end{table}
\par 
In continuation  Table~\ref{storage-episode} present the newly created episodes based on acceptable time difference considering value of $\tau$ is equal to 0.1. 
\begin{table}[h!]
\centering
\resizebox{\textwidth}{!}{%
\begin{tabular}{@{}lllllll@{}}
 \hline
 Instance &\multicolumn{2}{c}{Link} & Timestamp & Instance & Next & Utility \\
\cline{2-3}
 & ID & Weight & & & & \\
\hline
4300673472 & None & 0 & 2018-12-09 10:09:08. 532000 & 4300679876 & 4300673572 & 1 \\
4300673572 & None & 0 & 2018-12-09 10:03:09. 452000 & 4300679234 & None & 1 \\
\hline
\end{tabular}
}
\caption{New instance of episode after encoding}
\label{storage-episode}
\end{table}
\par
Broadening the knowledge infrastructure has been designed to work implicitly. Where system learns through experience to upscale the knowledge acquisition. The episodic memory has been developed  in such a way that the system learns over time as more and more information is fed to the system. 

\subsection{Learning Experience}
\begin{figure}[htb!]
\centering
\includegraphics[scale=0.5]{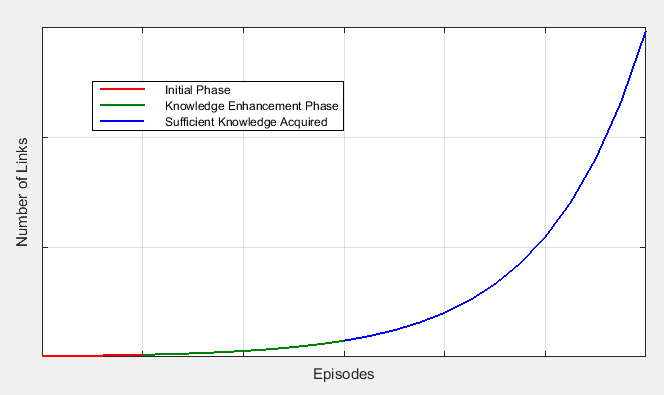}
\caption{Learning shown as the variation of Total Number of Links in system with the formation of Episodes}
\label{learning}
\end{figure}

The estimated enrichment of the episodic memory is represented graphically through Figure.~\ref{learning}. Initially the episodic memory is created and linked to each other only through chronological sequence as they rarely have anything in common. However later on, slowly but steadily the links start to rise as the system starts finding some knowledge in common. After sufficient accumulation of knowledge, there is a steep growth in links as for almost every new memory, similar memory can be found out in storage. 
Learning of such kind eliminate the redundant knowledge which significantly reduce the dense search space to a sparse search space. Which should improve the retrieval opportunity of a query.
\subsection{Retrieval Time}
Time to retrieve older memories will change with the accumulation of knowledge. Best Case will be observed when continuous knowledge on the same topic will be given.
Average Case will be observed when sufficient knowledge has been obtained and the topic on which knowledge is obtained is previously present in the memory.
Worst case scenario will be observed when the topic is not previously present, therefore it is required to search for the topic till the start of time.
\begin{figure}[h!]
\centering
\includegraphics[scale=0.4]{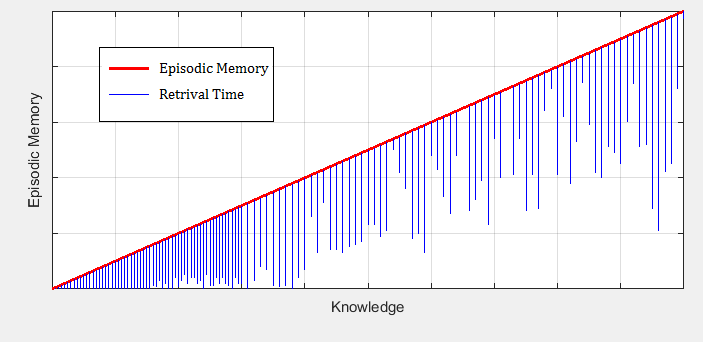}
\caption{Efficiency of the System in terms of Retrieval Time}
\label{retrieval}
\end{figure}
\par
A graphical representation of the memory retrieval rate with respect to knowledge acquired has been given in Figure~\ref{retrieval}. Here, retrieval time denotes the time taken in order to find the similar node. It is based on the observance that, if the knowledge network increases the system needs to go to the start of time less and less. Which is because the system only has to find the latest instance of topic and updates the links in the episodic memory accordingly.
\subsection{Question Answering}
In order to evaluate the working of episodic memory knowledge infrastructure, integration with a different application were carried out. The objective was to demonstrate the functioning of retrieval. 
\subsubsection{Analysis of the System}
Firstly, an analysis is done where the system is given a set of simple and complex sentences jumbled together. Now, a set of simple questions are fired on the simple text as well as the complex text. Similarly, a set of complex questions are also fired on on both the texts. Here, the questions are asked only for the cases where the answer is present in the text given to it previously. The results recorded are observed in Table.~\ref{Question Answering}.
\begin{table}[htb!]
\centering
\resizebox{0.5\textwidth}{!}{
\begin{tabular}{c c c } 
 \hline
 & Simple Question & Complex Question \\ 
 & (20) & (20)\\
 \hline
Simple sentence & 19 & 11\\
Complex Sentence & 9 & 3\\
\hline
\end{tabular}
}
\caption{Results of Question Answer Implementation for different scenarios}
\label{Question Answering}
\end{table}
\subsubsection{Comparison with Cleverbot}
For demonstrating the capabilities of our system with respect to other artificial intelligence question answering machine, we compare our system with "Cleverbot"\cite{carpenter2015cleverbot}. Cleverbot is a very popular web application which was developed by Rollo Carpenter. The reason for selecting Cleverbot when there are so many question answering machines available is its unique feature of developing database by having conversation with people. During its launch it had 200 million conversations which now has increased to 265 million. When asked a question, Cleverbot tries and matches it to the exact phrase. If no exact phrase is found, it searches for keywords in input and then retrieves the best match from database. 
\begin{table}[htb!]
\centering
{
\begin{tabular}{c c c} 
 \hline
 & Episodic Memory & Cleverbot \\ 
 \hline
Known Topic & 75\% & 90\% \\
New Topic & 0\% & 95\%\\
\hline
\end{tabular}
}
\caption{Efficiency of Our System \& Cleverbot based on Correct Answers}
\label{Cleverbot}
\end{table}
\par
Therefore, to show the comparison between the two, knowledge related question were fed to our system. As shown in Table.~\ref{Cleverbot} the results were found to be comparable when our system was fed appropriate data. However, it could not answer any questions for new topic because it entirely depends upon its accumulated knowledge and cannot give answers to such questions. We can see that the Cleverbot performs better for any given topic. Although, in case of a known topic, keeping in mind the vast difference between the database of the two, the observations made were quite satisfactory.

\section{Conclusion}
\label{conclusion}
In this paper we have presented a psychologically plausible knowledge representation infrastructure to organize text data. The intuition was to have text as a knowledge. Formal structure of an artificial episodic memory and number of operators were defined. Wordnet was used to supersede the requirement of semantic memory. Proper functioning episodes and operators were examined. Finally evaluation of the knowledge structure were presented to establish the claim. Looking at ways to deal with topics not seen before is part of an ongoing research. 
\section*{References}

\bibliography{elsarticle-template.bib}

\end{document}